%% file: main.tex
\documentclass{article}
\usepackage{arxiv}
\usepackage[utf8]{inputenc} % allow utf-8 input
\usepackage[T1]{fontenc}    % use 8-bit T1 fonts
\usepackage[english]{babel}
\usepackage{hyperref}       % hyperlinks
\usepackage{url}            % simple URL typesetting
\usepackage{booktabs}       % professional-quality tables
\usepackage{amsmath}        % math environments and commands
\usepackage{amsfonts}       % blackboard math symbols
\usepackage{nicefrac}       % compact symbols for 1/2, etc.
\usepackage{microtype}      % microtypography
\usepackage{cleveref}       % smart cross-referencing
\usepackage{lipsum}         % Can be removed after putting your text content
\setlipsum{language=english}
\usepackage{graphicx}
\usepackage{natbib}
\usepackage{doi}

\title{FORCE-Bench: A Benchmark, Dataset, and Evaluation Harness for Agentic AI in Enterprise Finance}

\usepackage{authblk}

\newbox{\orcid}\sbox{\orcid}{\includegraphics[scale=0.06]{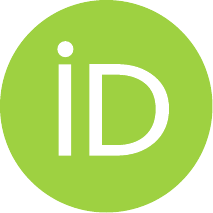}} 
\author[1]{%
	\href{https://orcid.org/0000-0002-0966-0254}{\usebox{\orcid}\hspace{1mm}Wolfgang M. Pauli % \thanks{\texttt{alias@microsoft.com}}
    }%
}
\author[1]{%
	\href{https://orcid.org/0009-0008-4206-9774}{\usebox{\orcid}\hspace{1mm}Sarah Panda % \thanks{\texttt{alias@microsoft.com}}
    }%
}
\author[1]{%
	\href{https://orcid.org/0000-0002-0654-1946}{\usebox{\orcid}\hspace{1mm}Kidus Admassu % \thanks{\texttt{alias@microsoft.com}}
    }%
}
\author[1]{%
	\href{https://orcid.org/0009-0005-7083-6852}{\usebox{\orcid}\hspace{1mm}Said Bleik % \thanks{\texttt{alias@microsoft.com}}
    }%
}
\author[1]{%
	\href{https://orcid.org/0000-0001-7445-3844}{\usebox{\orcid}\hspace{1mm}Ademola Okerinde % \thanks{\texttt{alias@microsoft.com}}
    }%
}
\author[1]{%
	\href{https://orcid.org/0009-0007-2688-3736}{\usebox{\orcid}\hspace{1mm}Jeremy Reynolds % \thanks{\texttt{alias@microsoft.com}}
    }%
}
\affil[1]{Microsoft Corporation, 1 Microsoft Way, Redmond, WA 98052}

\renewcommand{\headeright}{Technical Report}
\renewcommand{\undertitle}{Technical Report}
\renewcommand{\shorttitle}{FORCE-Bench}

\hypersetup{
colorlinks=false,
pdfborder={0 0 1},
citebordercolor={0 1 0},
linkbordercolor={1 1 1},
urlbordercolor={1 1 1},
filebordercolor={1 1 1},
pdftitle={FORCE-Bench},
pdfsubject={agentic AI, Finance, Enterprise Resource Planning},
pdfauthor={Wolfgang M. Pauli, PhD; Sarah Panda, PhD; Kidus Admassu, PhD; Said Bleik, PhD; Ademola Okerinde, PhD; Jeremy Reynolds, PhD},
pdfkeywords={agentic AI, Finance, Enterprise Resource Planning},
}

\begin{document}
\maketitle

\begin{abstract}
Recent advances in large language models have accelerated deployment of agentic systems in operational finance.
Existing benchmarks emphasize measuring general capabilities, instruction following, or safety, but few directly address the operational finance workflows that agentic systems are now being deployed to automate.
Finance professionals require agents to not only provide factually sound and properly grounded information, but also ensure that this information is verifiable and consistently adheres to rules and constraints of the operational finance domain.
We introduce FORCE-Bench, which contains 251 expert-annotated queries and evaluates responses using a rubric-based framework calibrated to the requirements of the operational finance domain, across eight dimensions: accuracy, citations, clarity, depth, groundedness, recency, relevance, and structure.
FORCE-Bench assesses agentic systems on three task types: financial obligation research (querying ERP systems for accounts receivable and payable data), financial entity performance research (answering time-bound questions from public filings and market data), and business brief generation (synthesising multi-source company intelligence reports).
To reflect real deployment conditions, we evaluate our purpose-built agent, as well as the general-purpose agentic systems, under common tool access and latency-bounded settings.
Results show that general-purpose agentic systems do not consistently meet finance-domain quality requirements under operational constraints, while the purpose-built Finance Agent for Microsoft 365 Copilot is more reliable across dimensions.
We release the dataset, rubrics, harness, and analysis code as open-source to support reproducible comparison and adaptation to other enterprise finance environments.
\end{abstract}

% keywords can be removed
\keywords{Agentic AI, Finance, Enterprise Resource Planning, Benchmark, Evaluation}

\section{Introduction}

There is growing excitement around the potential of AI agents - systems that autonomously reason, retrieve information, and act through tools to complete multi-step tasks - to unlock new levels of automation across industries.
Finance is among the most promising domains for such systems, yet realising that potential requires that agent performance be measured in a manner appropriate to the domain, meeting the expectations of finance professionals.
General-purpose LLM benchmarks - such as MMLU~\citep{hendrycks2020measuring}, BIG-Bench~\citep{srivastava2023beyond}, MT-Bench~\citep{zheng2023judging}, and TruthfulQA~\citep{lin2022truthfulqa} - were designed to assess broad model capabilities and alignment, not whether an agent can reliably serve a finance professional within the constraints and expectations of real-world operations.
Existing agent benchmarks, including AgentBench~\citep{liu2024agentbench}, $\tau$-Bench~\citep{yao2024taubench}, ReAct~\citep{yao2022react}, and ToolBench~\citep{yu2026toolbench}, establish important foundations for evaluating interactive, tool-using systems but target general-purpose or retail-domain scenarios rather than the workflows, data sources, and professional standards of the finance domain.
We introduce \textbf{FORCE-Bench} (\textbf{F}inancial \textbf{O}bligations, \textbf{R}esearch, and \textbf{C}omposition \textbf{E}valuation), a workflow-oriented evaluation framework designed specifically for AI agents operating in operational finance.
The benchmark provides: (i)~input queries drawn from realistic finance scenarios, (ii)~expert-annotated ground-truth responses, (iii)~a rubric-based metric framework with per-dimension scoring definitions, and (iv)~an evaluation harness that others can use to test their own agents against our dataset, or their own datasets.

Importantly, the evaluation harness is decoupled from any specific agent implementation: practitioners may apply the evaluation rubric to their own datasets, adjust the agent harness configuration, or modify hyperparameters to reflect their own evaluation or deployment requirements.

The benchmark spans 251 expert-annotated queries across three finance task types: \emph{financial obligation research} (querying internal ERP systems for accounts receivable and payable data), \emph{financial entity performance research} (answering time-bound questions about publicly traded companies from filings and market data), and \emph{business brief generation} (synthesising multi-source structured company intelligence reports).
Rather than measuring success as a binary pass/fail rate - as in $\tau$-Bench~\citep{yao2024taubench} and related work - we evaluate responses along eight dimensions grounded in the expectations of finance professionals: accuracy, citations, clarity, depth, groundedness, recency, relevance, and structure.
Each dimension is operationalised through carefully curated assertions, enabling a granular understanding of where a system excels and where it falls short.
This nuanced view is essential for decision-makers considering whether to deploy an agentic system in operational finance workflows: it is not enough to know that a system succeeds or fails - one must understand \emph{why} and \emph{along which dimensions}.

The remainder of this paper is organised as follows.
Section~\ref{sec:related work} surveys related work on LLM and agent benchmarks.
Section~\ref{sec:methods} describes the three pillars of the benchmark: the query dataset (Section~\ref{sec:dataset}), the rubric-based evaluation framework (Section~\ref{sec:evaluation}), and the agentic harnesses and statistical methodology used to compare them (Section~\ref{sec:agentic_harnesses}).
We then present experimental results, followed by concluding remarks.

We hope that FORCE-Bench will serve as a useful resource for researchers and practitioners developing and evaluating AI agents for the finance domain.
The dataset, evaluation rubrics, harness implementations, and analysis scripts are publicly available at \url{https://github.com/microsoft/FinanceBenchmark}.

\section{Related Work}
\label{sec:related work}

\subsection{LLM Benchmarks}

Existing LLM benchmarks evaluate model capabilities across diverse dimensions, yet few address the domain-specific trustworthiness and operational reliability required for enterprise deployment.
Most focus on static knowledge, general instruction-following, or safety in isolation, whereas our setting emphasizes consistent rule adherence, grounded information retrieval, and source-backed explanations for finance operations.
We organize prior work into four categories:

\paragraph{General Capability Benchmarks.} Benchmarks like MMLU \cite{hendrycks2020measuring}, BIG-Bench \cite{srivastava2023beyond}, MATH \cite{hendrycks2021measuring}, and GSM8K \cite{cobbe2021training} assess fundamental model knowledge and reasoning across broad domains.
These primarily evaluate single-turn, task-level capabilities rather than end-to-end interactive task completion with domain constraints and tool-grounded evidence requirements.
Our work extends beyond isolated capability assessment to measure whether models can reliably complete realistic finance workflows while adhering to domain rules and citing authoritative sources.

\paragraph{Instruction Following \& Quality.} Benchmarks such as MT-Bench \cite{zheng2023judging} and AlpacaEval \cite{li2023alpacaeval} evaluate models' ability to follow complex instructions and produce high-quality outputs.
While these capture general instruction adherence, they do not systematically measure domain-specific quality attributes (e.g., proper financial citations, accurate calculation provenance, regulatory compliance).
Our benchmark incorporates these professional quality dimensions, evaluating whether agents can meet the specific standards finance professionals require for trustworthy decision support.

\paragraph{Factuality \& Truthfulness.} TruthfulQA \cite{lin2022truthfulqa}, FactKG \cite{kim2023factkg}, and reading comprehension benchmarks like SQuAD \cite{rajpurkar2016squad} evaluate related aspects of factual reliability, including truthfulness under misconception-prone prompts, fact verification, and evidence-based question answering.
In contrast to these benchmarks, our benchmark evaluates agents' ability to maintain groundedness and accuracy throughout complex workflows with multiple data retrieval steps and tool call choices.

\paragraph{Safety \& Alignment.} Safety-focused evaluations and alignment work, including SafetyBench \cite{zhang2024safetybench}, ToxiGen \cite{hartvigsen2022toxigen}, and Constitutional AI \cite{bai2022constitutional}, assess harmful outputs and alignment with human values.
While essential for general-purpose models, these do not address the specific trust and reliability concerns in enterprise settings; e.g., whether agents consistently follow domain policies, correctly interpret regulatory constraints, and transparently acknowledge data limitations.
Our benchmark targets these domain-operational trust dimensions alongside safety.

\subsection{Finance-focused Benchmarks}

Finance-focused evaluation efforts have expanded rapidly, but they emphasize different slices of the problem space than our workflow-oriented setting.
FinanceBench \citep{islam2023financebench} established an important baseline for open-book financial QA with 10,231 public-company questions and evidence strings, showing that strong frontier models still struggle on finance-domain factuality.
However, its design centers on single-turn QA over public documents rather than ERP-grounded enterprise workflows or multi-source report synthesis.

BloombergGPT \citep{wu2023bloomberggpt} advanced domain-specialized model evaluation by combining open financial tasks with internal enterprise-aligned benchmarks, and showed that mixed-domain pretraining can improve finance performance without sacrificing general capabilities.
Yet this evaluation setup is tied to a model-development study and partially proprietary benchmark suites, limiting direct reuse as an open, standardized agent benchmark.

FinBen \citep{xie2024finben} broadened financial LLM assessment to 36 datasets and 24 tasks spanning extraction, QA, generation, forecasting, and decision-making.
This breadth is highly valuable for landscape-level model comparison, but it prioritizes broad task coverage over enterprise operational realism and does not focus on ERP-grounded agent behavior under deployment constraints.

\subsection{Agent Benchmarks}

Recent benchmarks have begun to address agent evaluation, focusing on interactive task completion in diverse environments.
Most emphasize task success or capability metrics (e.g., function calling, reasoning, navigation), while our setting adds professional quality dimensions for finance workflows: accuracy, citations, clarity, depth, groundedness, recency, relevance, and structure.
We categorize relevant agent benchmarks as follows:

\paragraph{Interactive Environment Benchmarks.} AgentBench \cite{liu2024agentbench}, WebShop \cite{yao2022webshop}, and ScienceWorld \cite{wang2022scienceworld} provide diverse simulation environments for agent evaluation.
AgentBench spans 8 distinct environments (web browsing, database query, file I/O, etc.); WebShop simulates e-commerce with 1.18 million real-world products; ScienceWorld tests scientific reasoning in interactive scenarios.
While valuable, these benchmarks primarily measure task success rates without evaluating grounding, citations, or compliance with domain-specific policies.

\paragraph{Tool-Use and Function Calling.} Gorilla \cite{patil2024gorilla}, ReAct \cite{yao2022react}, and ToolBench \cite{yu2026toolbench} focus on agents' ability to correctly use tools, call APIs, and reason about function selection.
Gorilla introduces APIBench for assessing function-calling accuracy; ReAct proposes a synergistic reasoning-acting framework; ToolBench benchmarks tool-use in real-world scenarios.
These address a critical capability but primarily focus on function-calling accuracy rather than evaluating response quality across dimensions relevant to professional applications.

\paragraph{Interactive Agent-User Benchmarks.} $\tau$-Bench \cite{yao2024taubench} emulates dynamic conversations between simulated users and agents with domain-specific tools and policy guidelines.
$\tau$-Bench evaluates agents' ability to interact naturally while adhering to rules in multi-turn scenarios.
This represents a closer alignment with our work, though $\tau$-Bench evaluates agent success through database state verification rather than through professional quality dimensions like citations, clarity, depth, recency, relevance, and structure.

\paragraph{Finance Agent Benchmarks.} Recent work has begun to benchmark agentic systems directly on finance tasks.
The Finance Agent Benchmark \citep{bigeard2025financeagentbenchmark} evaluates LLM agents on 537 expert-authored real-world financial research questions with tool access (e.g., search and EDGAR), providing a strong public-finance comparator.
FinAgentBench \citep{choi2025finagentbench} focuses on agentic retrieval in financial QA, separating document-type selection from passage selection to stress-test multi-step retrieval reasoning.
Together, these efforts materially advance finance-agent evaluation, but they primarily target public-company research and retrieval settings; in contrast, FORCE-Bench additionally evaluates enterprise ERP-grounded obligations workflows and structured business-brief synthesis under common latency constraints.

\subsection{Benchmark Landscape Summary and Design Goals}

Taken together, prior benchmark families establish a strong foundation but cover different layers of the evaluation problem.
In short, general LLM benchmarks characterize broad reasoning and instruction-following ability, agent benchmarks test interactive tool use and multi-step execution, and
finance-focused benchmarks improve domain relevance, especially for public-document question answering.
This progression clarifies what is already well measured and where enterprise finance deployment needs remain under-specified.

Three gaps motivate FORCE-Bench.
First, existing benchmarks rarely evaluate end-to-end operational finance workflows that combine internal ERP records with external financial sources.
Second, many evaluations emphasize task completion or retrieval success, but provide limited visibility into professional-quality dimensions such as citation quality, groundedness, clarity, structure, and depth.
Third, benchmark settings often abstract away deployment constraints (for example, latency budgets and shared tool-access conditions) that materially affect practical usefulness.

Accordingly, FORCE-Bench is designed around three goals: (i) realistic workflow coverage across obligations research, entity-performance research, and business-brief generation; (ii) diagnostic, rubric-based multidimensional scoring aligned with finance-professional expectations; and (iii) reproducible, implementation-agnostic evaluation under operationally relevant constraints.
These goals directly inform the dataset construction, evaluation framework, and harness methodology described in the following section.

\section{Methods}
\label{sec:methods}

This section describes the three pillars of the benchmark: the query dataset (Section~\ref{sec:dataset}), the rubric-based evaluation framework (Section~\ref{sec:evaluation}), and the statistical methodology used to compare agentic harnesses (Section~\ref{sec:stats}).
Dataset, evaluation rubrics, harness implementations, and analysis scripts are publicly available at \url{https://github.com/microsoft/FinanceBenchmark}.

\subsection{Dataset}
\label{sec:dataset}

The benchmark dataset comprises 251 queries spanning three task types, each testing a distinct finance-agent capability:

\paragraph{Financial obligation research.} This task type comprises internal accounts receivable (AR) and payable (AP) database queries grounded in a synthetic Dynamics 365 Finance instance (see below for details).
Queries cover customer balance inquiries, aged debt analysis, invoice details, payment history, vendor terms, and collection activities.
The dataset contains 100 financial obligation queries: 72 AR-focused and 28 AP-focused.
Representative scenarios include \emph{Aged Balance}, \emph{Cash Collections}, \emph{Cash Discounts}, \emph{Collections}, \emph{Collections Tasks}, \emph{Credit Limit}, \emph{Credit Notes}, \emph{Credit Rating}, \emph{Customer Setup}, \emph{Discounts}, \emph{Dispute}, \emph{Invoicing History}, \emph{Outstanding Balance}, \emph{Payment History}, \emph{Sales Orders}, \emph{AP Invoices}, \emph{AP Payments}, \emph{AP Purchase Orders}, \emph{Vendor Balance}.

\emph{Example query:} ``What is the outstanding receivables balance for Birch Company as of March 2, 2026?''

\emph{Expected answer:} A structured response identifying the AR balance amount, customer identifier, and relevant aging or payment status, grounded in ERP data with a reference to the ERP system or record.

\paragraph{Financial entity performance research.} Public-company financial performance queries drawn from earnings reports, SEC filings, and financial data providers.
This set contains 126 queries covering topics including earnings per share, debt-to-equity ratios, revenue figures, cash flows, capital expenditures, inventory turnover, and ESG metrics.
Queries are scoped to specific fiscal periods and require agents to locate and cite current public data.

\emph{Example query:} ``For our credit evaluation of Caterpillar, what is their long-term debt-to-equity ratio (3-year average) as of September 2025? I need this to inform our leverage assessment.''

\emph{Expected answer:} ``Caterpillar's long-term debt-to-equity ratio (3-year average) as of September 2025 is 1.42 (or 142\%).
This metric represents the relationship between Caterpillar's long-term debt and equity, measuring the company's long-term capital structure.
Source: Caterpillar Inc.\ Quarterly Earnings \& Form 10-K Filings.''

\paragraph{Business brief generation.} Requests for structured company profiles synthesizing financial data, business context, and operational details.
The dataset includes 25 business brief queries.
Responses are multi-section reports covering company overview, financials, lines of business, competitive positioning, and strategic moves.
These tasks evaluate synthesis and depth across heterogeneous data sources.

\emph{Example query:} ``Business Brief report of Apple Inc.''

\emph{Expected answer:} A comprehensive structured document with sections on company overview, financial performance (income statement, balance sheet, ratios), lines of business/segments, competitive positioning relative to peers, recent strategic moves, and key metrics-all grounded in cited sources.

\subsection{Evaluation Overview}
\label{sec:evaluation}

Responses are evaluated against a rubric-based framework grounded in finance-domain expectations.
Each question is annotated with a set of evaluation dimensions (tags), each with multiple assertions.
An assertion is a testable claim about what a finance professional would expect or value in a response-e.g., ``the response grounds financial figures in cited sources'' or ``the response defines uncommon acronyms on first use.''  These assertions are designed to be clear and diagnostic, supporting consistent evaluation across responses.

Performance is measured across eight dimensions: accuracy, citations, clarity, depth, groundedness, recency, relevance, and structure.  Only a subset of these eight dimensions is evaluated for all three task types: citations, clarity, depth, groundedness, relevance, structure.  We only include the dimension recency for financial entity performance research and business brief generation, while recency is omitted for financial obligation research, because this dimension doesn't readily apply when querying a system of record (Dynamics 365 for Finance).  On the other hand, financial obligation research uniquely includes accuracy, while we don't measure accuracy for the other two tasks, as ground truth can sometimes be subjective and rapidly evolving over time.  We instead focus on groundedness for those two task types.

Even for the common set of dimensions, only some assertions are shared across task types, while we also added task-specific assertions tailored to the task type's data sources and complexity.  For instance, for business brief generation we include section-specific depth assertions (e.g., ``response must include a Company Overview section'') whereas financial entity performance research includes topic-specific depth assertions (e.g., ``response includes peer comparisons'' for queries on competitive positioning).

Evaluation is performed by an LLM judge, implemented using DSPy~\citep{khattab2023dspy}, that scores each assertion on a continuous 0.0--1.0 scale, where intermediate values (0.25, 0.5, 0.75) represent partial compliance. A dimension's score is the mean of its assertion scores.

As default, we used OpenAI GPT-5.2 as the judge model consistently across all evaluation runs.  For a comparison between two different judges, see Appendix~\ref{app:judge_bias}.

\subsection{Metric Dimensions and Rubrics}
\label{sec:dimensions}

The rubric design process was iterative and collaborative, involving multiple stakeholders to ensure that the metrics were both relevant and actionable.
The initial list of assertions was informed by input from internal subject matter experts and qualitative survey feedback from external finance professionals.
The assertions were then calibrated against real model outputs, leading to the removal of saturated assertions and the revision of ambiguous assertions.
Throughout this process, the focus remained on finance-specific dimensions to ensure the rubric's relevance to the domain. The resulting dimensions are described below.

\paragraph{Accuracy.} Correctness of factual claims against ground-truth assertions.
This metric is evaluated only for financial obligation research, where ERP-sourced figures (customer balances, invoice amounts, and payment status) can be verified against the system of record.
It is not applied to financial entity performance research or business brief tasks, where ground truth is distributed across sources and subject to temporal drift or definitional variation. For instance, a P/E ratio may differ across sources depending on whether a trailing-twelve-month or forward estimate is used; groundedness serves as the primary factual-grounding signal for those task types instead.
Example assertion: \emph{``the response's key facts (customer balances, invoice amounts, payment status) match the ground truth as recorded in the ERP system.''}

\paragraph{Citations.} Presence and appropriateness of source references.
Financial responses must cite their data sources to support verification and professional decision-making.
For financial obligation research, citations reference ERP entities or system URIs---e.g., \emph{``response includes at least one URL linking to the ERP record or entity; absence of a URL is acceptable only if no records were retrieved.''}
For financial entity performance research, citations must name a traceable financial data provider---e.g., \emph{``response cites the data source by naming the provider (e.g., Bloomberg, S\&P Capital IQ, or company filings).''}
For business briefs, each section must be grounded in a named source or inline hyperlink---e.g., \emph{``response cites data sources by naming the provider (e.g., Bloomberg, S\&P Capital IQ, company filings) or supplying a traceable URL.''}.

\paragraph{Clarity.} Quality of explanation and definitional precision.
Responses must define unusual acronyms on first use (standard finance abbreviations such as GAAP, EBITDA, and DSO are assumed to be familiar to finance professionals) and maintain consistent terminology throughout.
Responses should organize information hierarchically and avoid vague qualifiers (e.g., ``strong'' or ``significant'' without quantification).

\paragraph{Depth.} Sufficiency and breadth of context.
For financial obligation research, depth means covering all relevant dimensions of a query (e.g., aging buckets, terms, and payment history if requested).
For financial entity performance research, depth requires claim-level precision---e.g., \emph{``every claim maps to a number, date, or named source; approximate qualifiers are used only if precise values are unavailable''}---and comparative context (3-year trends, peer benchmarks, and period-over-period change) where available.
For business briefs, depth is assessed section by section---e.g., \emph{``response includes a Company Overview section covering the company summary, leadership, founding, and locations''} and \emph{``response includes a Competitive/Peer Analysis section comparing peers on multiple attributes.''}.

\paragraph{Groundedness.} Factual claims traceable to retrieved source content.
The benchmark captures source snippets and tool outputs at inference time and evaluates whether response assertions are supported by, or reasonably inferred from, those sources.
Sources are drawn from ERP system outputs (financial obligation research), web-retrieved content and financial databases (financial entity performance research), and multi-source synthesis (business briefs).

\paragraph{Recency.} Use of appropriately current data.
Responses must anchor time-sensitive data to specific dates or periods (e.g., ``as of Q3 2025'' rather than ``currently'') and prioritize the most recent available figures.
This metric is evaluated for financial entity performance research and business briefs, where data currency directly affects decision quality. It is not measured for financial obligation research, where the ERP snapshot date is fixed at inference time.
For financial entity performance research, assertions require explicit date anchoring---e.g., \emph{``response anchors time-sensitive data to a specific date or period rather than using relative terms such as currently or recently.''}.
For business briefs, recency is assessed in terms of data-intent alignment---e.g., \emph{``response aligns the recency of data with the user's intent, such as the latest quote for a current-price query versus historical figures for trend analysis.''}.

\paragraph{Relevance.} Directly addressing the user's query without tangential information.
A relevant response answers what was asked, avoids unrelated digressions, and communicates data limitations honestly when requested information is unavailable.
For financial obligation research, relevance means focusing on the specific legal entity, customer, or time period specified in the query; for financial entity performance research, it means avoiding unrelated company or peer information; for business briefs, it means staying within company-specific scope.

\paragraph{Structure.} Organization and formatting quality.
Responses should present the most critical information first (answer before detail), use clear hierarchies (headings, sections, and tables) where appropriate, and maintain consistent formatting across comparable data (e.g., currency, date formats, and table structure).
For financial obligation research, structure assertions additionally enforce a tabular layout for responses with multiple records---e.g., \emph{``if two or more records are returned, the response presents them in a structured table with clear column headers and consistent formatting (e.g., currency, dates, customer IDs).''}.

\subsubsection{Dataset Curation}

Curation spanned multiple stages: task design, query generation, grounding context assembly, and iterative refinement.
We began by defining representative capabilities for each of the three task types - business brief generation, financial entity performance research, and financial obligation research - and constructed query sets reflecting real-world user scenarios.
We continuously updated the datasets based on evaluation results, expanding coverage to include challenging cases, edge conditions, and observed failure modes.

\paragraph{Business brief generation.} The business brief generation dataset consisted of prompts that triggered the generation of structured, multi-section reports over real-world entities, each paired with curated grounding context from financial documents and retrieval pipelines, enabling evaluation of synthesis quality rather than exact answer matching.
We incorporated an SME-in-the-loop process in which model-generated responses were compared, filtered, and refined based on practitioner preferences - yielding what we term \emph{semi-golden} reference outputs: responses that reflect practitioner expectations around structure, clarity, depth, and source attribution without assuming a single immutable answer.
The dataset emphasized realistic consumption patterns of business briefs, where usefulness, traceability of claims, and organization of insights are critical for downstream decision-making.

\paragraph{Financial entity performance research.} The financial entity performance research dataset spanned a diverse set of financial reasoning tasks, including performance analysis, market dynamics, and risk interpretation.
We generated and filtered queries to ensure topical coverage and relevance, providing grounding context through associated financial data and retrieval results.
Because answers in this domain can evolve over time or depend on interpretation, we prioritized representative coverage and high-quality reference responses over fixed canonical answers, with SME feedback playing a central role in refining both queries and responses.

\paragraph{Financial obligation research.} The financial obligation research dataset was constructed over a realistic Finance and Operations (FnO) environment populated with synthetic but structurally faithful enterprise data, including customers, vendors, and transactional records.
We focused curation on the breadth and complexity of ERP workflows, incorporating scenarios such as multi-entity queries, cross-module reasoning, ambiguity in entity resolution, and edge cases.
Queries were paired with grounded answers derived directly from the underlying data environment, enabling reproducible and consistent evaluation; where needed, prompts included explicit clauses such as \emph{``as of''} dates for aged balance questions to anchor responses to the existing data state and avoid triggering processes, such as aging snapshot generation, that could alter the ground truth.
We iteratively expanded the dataset to address coverage gaps and stress-test system capabilities. The FnO environment included data describing 1000 vendors and 1000 customers, and included over 3400 AR and AP journal entries (each).

\subsection{Agentic Harnesses}
\label{sec:agentic_harnesses}

The benchmark is evaluated across three agentic harnesses: a purpose-built Finance Agent in Microsoft 365 Copilot, and two general-purpose harnesses - the Anthropic Claude CLI and the OpenAI Responses API.
The latter two were included not to replicate the Finance Agent, but to measure the out-of-the-box experience a user would obtain by deploying a general-purpose agentic system on finance queries.
Both receive the same shared system instructions and tool access, and differ from a vanilla deployment only in three minimal augmentations: (i) they are told the benchmark comprises three task types, (ii) they are explicitly permitted to use web search and URL fetching, and (iii) they are instructed to direct all accounts receivable and payable queries to the ERP MCP server rather than relying on general web knowledge.
Full configuration details are available in the public repository: \url{https://github.com/microsoft/FinanceBenchmark}.

For the main experiments, all three harnesses were run with the same per-question time budget applied to the Finance Agent: 60\,s for financial obligation research and financial entity performance research, and 300\,s for business brief generation.
This asymmetry reflects different user-experience expectations for each task: the first two task types are designed for chat-like, near-real-time retrieval, while business brief generation involves multi-source synthesis and is afforded proportionally more time.
General-purpose harnesses are evaluated at \texttt{reasoning\_effort=low} (OpenAI) and standard mode without extended thinking (Claude CLI) throughout the main experiments.
Enabling higher reasoning effort would cause the 60\,s timeout to bind even more frequently, further degrading performance; per-provider latency distributions under default reasoning settings are reported in Appendix~\ref{app:latency}.

\paragraph{Claude CLI.}
The Claude CLI harness invokes Anthropic's \texttt{claude} command-line tool as an asynchronous subprocess, with output streamed as newline-delimited JSON events.
This harness is agentic in the sense that the CLI manages a multi-turn reasoning-and-tool-use loop autonomously: at each turn the model may call one or more tools, inspect their outputs, and decide whether to continue or produce a final response, up to a maximum of \texttt{max\_turns\,=\,20} turns.
The models used were \texttt{claude-opus-4-7} and \texttt{claude-haiku-4-5}; an optional \texttt{effort} parameter controls the extended thinking budget (low / medium / high).
Available tools are web search (via the built-in \texttt{WebSearch} permission) and the ERP MCP server (provided via a filtered \texttt{--mcp-config} pointing to a single MCP server).
File-system tools (\texttt{Read}, \texttt{Write}, \texttt{Edit}, \texttt{Bash}, \texttt{Grep}, \texttt{Glob}) and skill-management tools unrelated to benchmarking are explicitly blocked.

\paragraph{OpenAI Responses API.}
The OpenAI harness calls the Responses API (\texttt{client.responses.create}), which differs from the Chat Completions API in that the agentic loop - tool selection, execution, and iteration - is managed server-side rather than by the caller.
The model issues tool calls autonomously until it produces a final response or reaches \texttt{max\_tool\_calls\,=\,20}.
The model used is \texttt{gpt-5.5}; an optional \texttt{reasoning\_effort} parameter (low / medium / high) controls the reasoning token budget for reasoning models.
Two tool types are registered natively: an MCP server entry pointing to the ERP endpoint and the built-in \texttt{web\_search\_preview} tool.
\label{sec:stats}

\subsubsection{Comparative Analysis of Agentic Harnesses and Inference Runs}

To statistically compare performance across different inference runs or across different agentic harnesses (Finance Agent, Claude CLI, and OpenAI Responses API), we employ Linear Mixed Models (LMM) that account for the repeated-measures structure of the data: the same questions are evaluated under each provider, so observations within a question are not independent across providers.

\paragraph{Model Specification.}
For each evaluation metric, we fit a mixed-effects model of the form:
\begin{equation}
\text{score} \sim C(\text{provider}) \times C(\text{task type}) + (1 \mid \text{question})
\end{equation}
where $C(\cdot)$ denotes categorical (fixed) effects.
The model includes a main effect for provider, a main effect for task type, and their interaction.
The random intercept grouped by question ID captures variation in baseline difficulty and idiosyncrasy across individual queries, accounting for the fact that some queries are inherently more challenging than others across all providers.
This random effect structure is essential: it ensures that provider comparisons are not confounded by differences in question difficulty, a critical consideration when evaluating agent performance across heterogeneous, realistic tasks.

\paragraph{Hypothesis Testing and Contrasts.}
We construct hypothesis tests for provider effects using two complementary contrast schemes \citep{judd1989data}.
First, we compute an \emph{equal-weighted marginal contrast} across task types, where each task type contributes equally regardless of the number of questions assigned to it.
This approach aligns with our primary aggregation strategy (weighted scores that average over task types with equal weight).
Specifically, let provider $A$ denote the reference and provider $B$ the comparator; the equal-weighted marginal contrast takes the form:
\begin{equation}
\delta_{A \to B} = \beta_{B} + \frac{1}{k} \sum_{j \neq j_0} \beta_{B \times j},
\end{equation}
where $\beta_{B}$ is the main effect of provider $B$ (relative to $A$), $\beta_{B \times j}$ are the provider-by-task-type interaction coefficients, $j_0$ is the reference task type, and $k$ is the total number of task types.
Second, we compute \emph{task-type-specific contrasts} from the same fitted model, allowing us to report provider effects within each task domain.

All models are estimated by maximum likelihood (ML) rather than restricted maximum likelihood (REML), which is required for valid fixed-effect inference from the estimated covariance matrix \citep{verbeke2000linear}.

For each fitted model we test provider contrasts using the Wald procedure \citep{verbeke2000linear}: given a contrast vector $\mathbf{c}$, the estimate and its standard error are
\begin{equation}
\hat{\delta} = \mathbf{c}^T \hat{\boldsymbol{\beta}}, \qquad
\widehat{\mathrm{se}}(\hat{\delta}) = \sqrt{\mathbf{c}^T \hat{\Sigma} \mathbf{c}},
\end{equation}
where $\hat{\boldsymbol{\beta}}$ is the vector of fixed-effect estimates and $\hat{\Sigma}$ is its estimated covariance matrix.
Under the null $\mathbf{c}^T \boldsymbol{\beta} = 0$, the statistic $z = \hat{\delta} / \widehat{\mathrm{se}}(\hat{\delta})$ is asymptotically standard normal, yielding two-tailed $p$-values $p = 2\,\Phi(-|z|)$.
We report $p$-values (rendered as significance markers in figures) alongside point estimates with 95\% bootstrap confidence intervals ($n = 1000$ resamples; error bars in figures and Table~\ref{tab:provider_scores}) to support direct interpretation of practical significance.
The LMM is used for significance testing only; the direction of each comparison is taken from the raw equal-weighted score difference to ensure consistency with the values shown in figures and Table~\ref{tab:provider_scores}.

\paragraph{Robustness.} We implement several safeguards to ensure reliable inference.
Models with non-convergence, zero variance in the outcome, or insufficient per-provider sample size (fewer than 5 observations) are flagged and excluded from reporting.
Missing data in score columns are dropped listwise within each metric, a reasonable assumption given that missing values in our LLM judge evaluations are rare and typically represent transient evaluation failures rather than systematic issues.
We do not apply explicit multiple-comparison corrections (e.g., Bonferroni) across metrics; instead, the focus on domain-specific metrics (clarity, groundedness, etc.) and effect-size reporting allows readers to assess evidence in context, a standard practice in applied ML benchmarking \citep{demsar2006statistical, dror2018hitchhiker}.

\section{Results}

\subsection{Overall Performance}

\begin{figure}[ht]
  \centering
  \includegraphics[width=\linewidth]{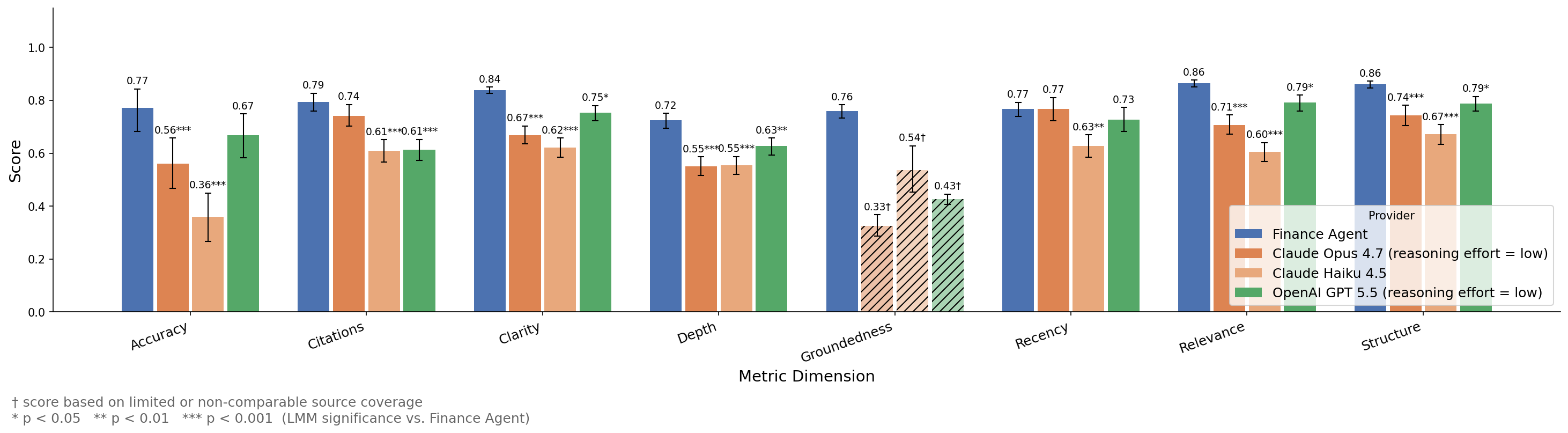}
  \caption{Finance Agent Benchmark scores across eight evaluation dimensions for Finance Agent (purpose-built), Claude Opus~4.7 and Claude Haiku~4.5 (Anthropic Claude CLI, reasoning effort = low), and GPT~5.5 (OpenAI Responses API, reasoning effort = low).
Error bars show 95\% bootstrap confidence intervals ($n=1000$ resamples); stars indicate statistical significance of the difference relative to Finance Agent ($^{*}p<0.05$, $^{**}p<0.01$, $^{***}p<0.001$, LMM).
Groundedness scores for general-purpose systems are shown with hatching ($\dagger$), reflecting limited or non-comparable source coverage.
Finance Agent leads across all dimensions, demonstrating the advantage of a purpose-built agent over general-purpose agentic runtimes on finance-domain evaluation criteria.}
  \label{fig:results}
\end{figure}

Figure~\ref{fig:results} and Table \ref{tab:provider_scores} show benchmark scores across all eight evaluation dimensions for the Finance Agent, Claude Opus~4.7 (Anthropic Claude CLI), Claude Haiku~4.5 (Anthropic Claude CLI), and GPT~5.5 (OpenAI Responses API).
The Finance Agent leads across all dimensions, with statistically significant margins on most metrics: accuracy of 0.77 compared to 0.67, 0.40, and 0.56 for GPT~5.5, Claude Haiku~4.5, and Claude Opus~4.7, respectively, and similar advantages in citations, clarity, depth, relevance, and structure.

\input{provider_scores_table.tex}

\subsection{Performance by Task Type}

\begin{figure}[t]
  \centering
  \includegraphics[width=\linewidth]{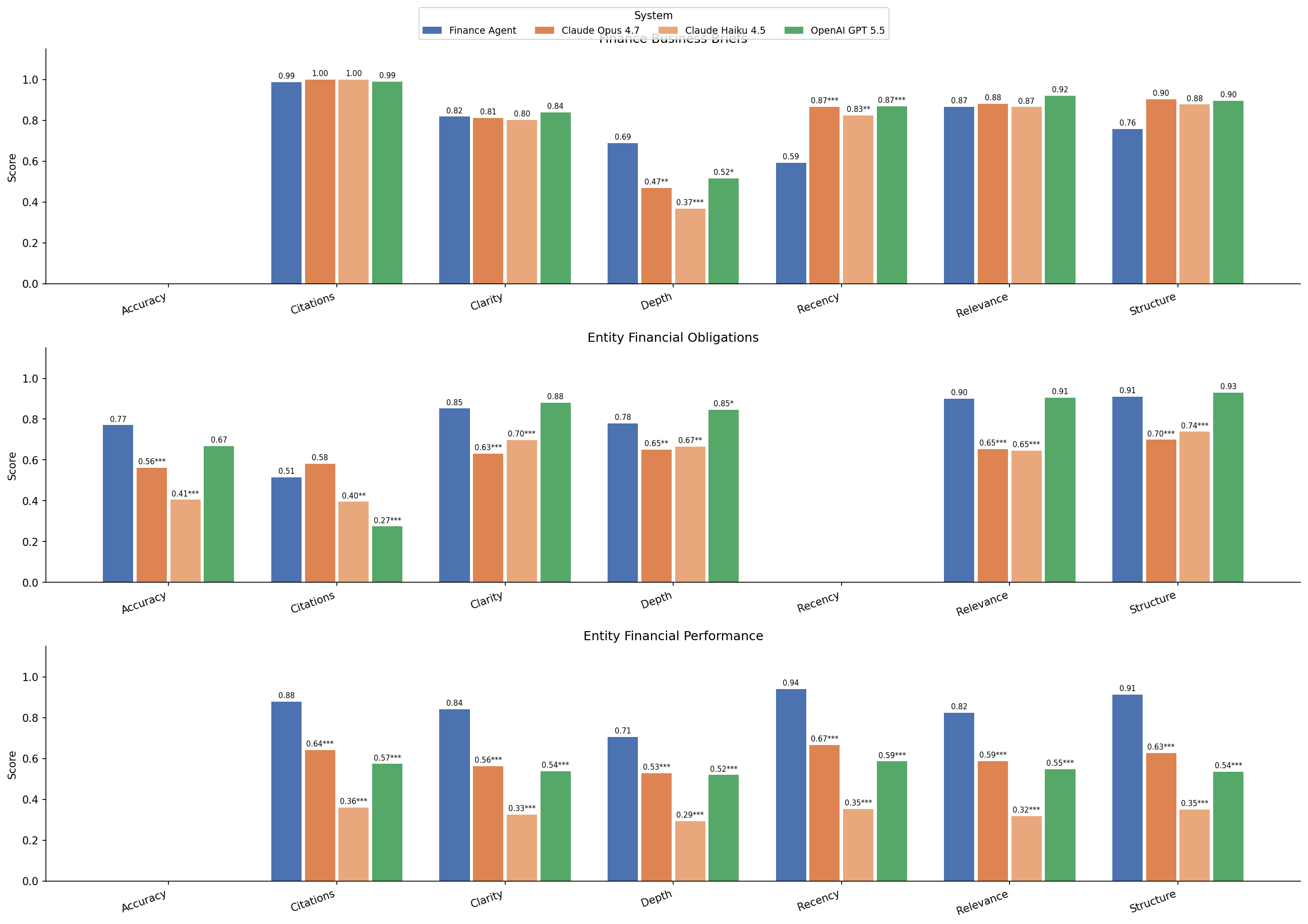}
  \caption{Per-task-type benchmark scores for all four evaluated systems across the applicable evaluation dimensions.
  \textbf{Top:} Finance Business Brief generation (\emph{business\_brief}); accuracy is not measured for this task type.
  \textbf{Middle:} Financial Obligation Research (\emph{erp\_qa}); recency is not measured because the ERP snapshot date is fixed at inference time.
  \textbf{Bottom:} Financial Entity Performance Research (\emph{finance\_qa}); accuracy is not measured because ground truth is distributed and time-dependent.
  Groundedness scores for the Claude systems on the bottom two tasks are near zero due to the absence of captured source content from their retrieval pipelines, not necessarily a reflection of factual error.
  Significance markers indicate LMM-based contrasts against the Finance Agent as reference: $^{*}p<0.05$, $^{**}p<0.01$, $^{***}p<0.001$.}
  \label{fig:scores_by_plugin}
\end{figure}

Figure~\ref{fig:scores_by_plugin} breaks down evaluation scores across all three task types.
Metric coverage intentionally varies by task: accuracy is measured only for financial obligation research, where ground truth is authoritative and deterministic (ERP system of record); recency is omitted for the same task type, since the ERP snapshot date is fixed at inference time and temporal anchoring of data is not a meaningful quality signal.
Conversely, accuracy is not applied to financial entity performance research  or business brief generation, where correct answers are distributed across multiple sources and change over time.

\subsubsection{Finance Business Brief Generation}
All systems achieve near-perfect citation scores (0.99--1.00), confirming that multi-section report generation reliably surfaces source references.
The Finance Agent leads on depth (0.69 vs.\ 0.37--0.52, significant at $p < 0.05$ or better for all comparators).
Conversely, recency is the one dimension where the Finance Agent trails all comparators: Claude Haiku~4.5 (0.83), Claude Opus~4.7 (0.87), and GPT~5.5 (0.87) each score significantly higher ($p \leq 0.01$ or better), indicating that the comparator systems more consistently anchor time-sensitive figures to explicit dates and periods.
For citations, clarity, relevance, and structure, the Finance Agent is on par with the leading provider on each dimension, with no comparator showing a statistically decisive advantage.
Accuracy is not evaluated for business briefs.

\subsubsection{Financial Obligation Research}
The Finance Agent leads on accuracy (0.77 vs.\ 0.67 for GPT~5.5, 0.56 for Claude Opus~4.7, and 0.41 for Claude Haiku~4.5), it is on par with the leading general-purpose system for the other dimensions, with a different general-purpose system leading on each of the other dimensions.
This reflects its purpose-built integration with the Dynamics~365 Finance ERP: queries about customer balances, invoice aging, or payment status are answered directly from system-of-record data, enabling high factual accuracy and tight source grounding.

\begin{figure}[ht]
  \centering
  \includegraphics[width=0.9\linewidth]{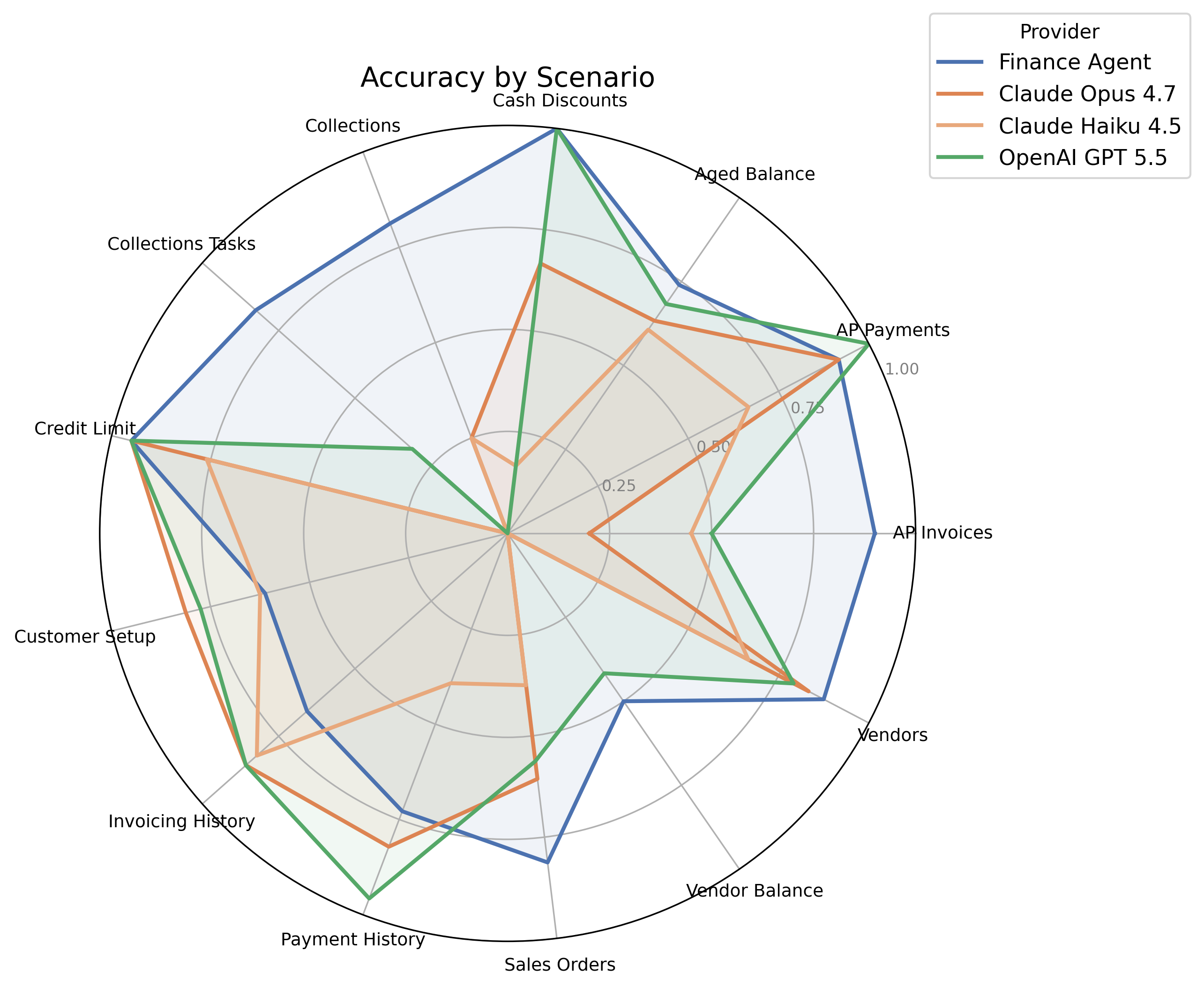}
  \caption{Finance Agent accuracy by ERP scenario for financial obligation research (\emph{erp\_qa}).
  Each spoke represents one scenario; the shaded area reflects mean accuracy across queries in that scenario.
  The Finance Agent's largest advantages occur in Collections Tasks, Collections, AP Invoices, and Sales Orders, where targeted ERP tool routing provides the greatest benefit over general-purpose retrieval.}
  \label{fig:radar_erp_scenario}
\end{figure}

Because financial obligation research is the only task type with deterministic ground truth, we can decompose accuracy further by ERP scenario.
Figure~\ref{fig:radar_erp_scenario} shows a radar chart of Finance Agent accuracy across all scenarios; the Finance Agent's advantage is most pronounced on four categories: \emph{Collections Tasks}, \emph{Collections}, \emph{AP Invoices}, and \emph{Sales Orders}.

\paragraph{Collections Tasks.} These queries ask the agent to navigate the collections worklist module: which open follow-up activities exist for a given customer, which customers appear on a collection agent's worklist for a specified date, and what open collection cases are outstanding across the organisation. Correct answers require resolving customer identifiers, reading task status fields, and returning structured lists of activities with dates and case states---a retrieval pattern the Finance Agent's ERP integration handles through purpose-built tool call instructions.

\paragraph{Collections.} These queries focus on the formal dunning letter process: which collection letters have been sent to a customer, when the most recent letter was dispatched, how many letters a given customer has received, and how many customers are unassigned to any collection pool. Success requires navigating the collection letter module and aggregating counts across entities---tasks that benefit from the Finance Agent's targeted ERP query routing.

\paragraph{AP Invoices.} The accounts-payable queries cover the invoice register: total value of unposted vendor invoices, number of invoices pending approval, amounts posted in a given period, and three-way match status between purchase orders, goods receipts, and invoices. These require access to AP-specific modules and correct interpretation of invoice lifecycle status (unposted, pending, matched), where the Finance Agent's dedicated AP tooling provides an advantage.

\paragraph{Sales Orders.} These queries cover open and historical sales orders by customer: the largest orders for a given customer, lists of open orders, return counts, and order-level details such as the salesperson who placed the order. Although sales orders are primarily a revenue-cycle construct rather than an AR metric, they feed directly into receivables context, and the Finance Agent's broader ERP integration surfaces these records more reliably than general-purpose web search.

\subsubsection{Financial Entity Performance Research}
The Finance Agent leads on all measured dimensions: recency (0.94 vs.\ 0.35--0.67), structure (0.91 vs.\ 0.35--0.63), citations (0.88 vs.\ 0.36--0.64), clarity (0.84 vs.\ 0.33--0.56), relevance (0.82 vs.\ 0.32--0.59), and depth (0.71 vs.\ 0.29--0.53), with all differences significant at $p < 0.001$ against every comparator.
These gains reflect the agent's curated retrieval pipeline and system instructions that explicitly require date-anchored sourcing and structured presentation.

\subsection{Influence of Reasoning Effort and Timeout constraint}
\label{sec:timeout}

All general-purpose harness runs reported in this paper use fixed reasoning settings: the OpenAI Responses API was configured with \texttt{reasoning\_effort=low}, while the Claude CLI was invoked without an explicit \texttt{effort} parameter, corresponding to standard mode without an extended thinking budget.
These choices reflect a low-latency deployment baseline; higher reasoning budgets would increase token cost and response time substantially.

To demonstrate the performance gains available when these constraints are relaxed, we ran an ablation in which GPT~5.5, Claude Haiku~4.5, and Claude Opus~4.7 were re-evaluated under a uniform 300\,s timeout across all task types, while raising reasoning effort from low to the default setting.
In practice, even with the raised reasoning effort, the 300\,s ceiling is rarely enforced: as shown in Appendix~\ref{app:latency}, p90 latencies under default reasoning settings reach at most 194\,s.
Enabling even higher reasoning budgets would likely further improve quality in the unconstrained regime, but would simultaneously amplify the degradation under the 60\,s budget by making individual tool-call sequences even more time-consuming.

\begin{figure}[t]
  \centering
  \includegraphics[width=\linewidth]{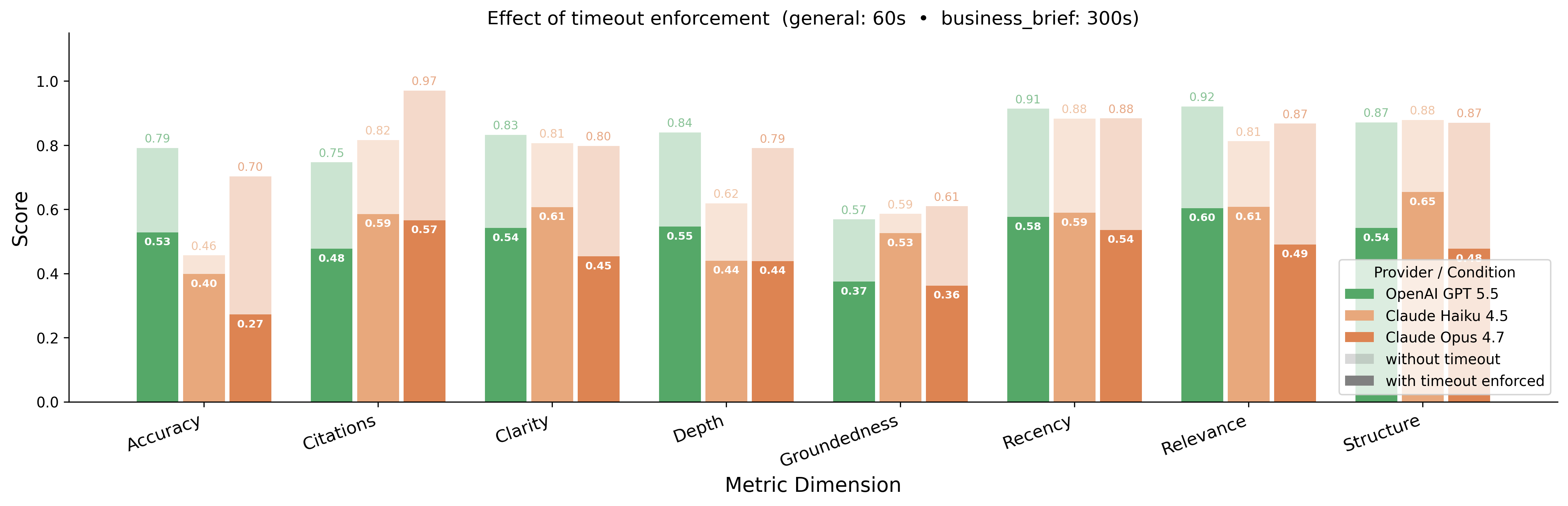}
  \caption{Effect of timeout enforcement on benchmark scores for GPT~5.5, Claude Haiku~4.5, and Claude Opus~4.7.
  Solid bars show scores under the enforced timeout (60\,s for financial obligation and entity performance research; 300\,s for business brief generation); faded bars show scores from unconstrained runs.
  All runs use the default reasoning settings for the OpenAI Responses API and for the Claude CLI.
  The Finance Agent is excluded from this analysis as it operates with a fixed effort setting.}
  \label{fig:timeout}
\end{figure}

Figure~\ref{fig:timeout} reports the results of this ablation, showing scores under the 300\,s unconstrained setting (faded bars) compared to the original 60\,s constrained runs (solid bars) with the increase in reasoning effort.

Enforcing the 60\,s timeout substantially degrades performance across most dimensions for all three providers.
The largest drops occur in accuracy (OpenAI: $-0.26$; Claude Haiku: $-0.24$), citations (Claude Opus: $-0.40$), depth (OpenAI: $-0.29$), and structure (OpenAI: $-0.33$), reflecting the fact that many financial obligation research queries require multiple sequential tool calls to retrieve ERP records or financial data before a complete answer can be assembled (per-provider tool-call distributions are reported in Appendix~\ref{app:tool_calls}).

An instructive reversal emerges within the Claude family on the accuracy dimension: under the 60\,s constraint, Claude Opus~4.7 (0.27) falls below Claude Haiku~4.5 (0.40), even though Opus outperforms Haiku when unconstrained.
This inversion is driven by latency: as shown in Appendix~\ref{app:latency}, Claude Opus produces longer, more deliberate tool-call sequences that require more wall-clock time to complete.
Under the 60\,s ceiling, Opus is timed out mid-retrieval more frequently than the faster Haiku, forcing an early termination that substantially reduces accuracy.
Haiku's smaller model footprint and faster response time allow it to complete most ERP lookups within the budget, avoiding this penalty.

Taken together, these results show that the choice of timeout is a meaningful hyper-parameter for practitioners deploying agentic systems on finance tasks: thresholds that are appropriate for latency-sensitive operational queries (60\,s) impose a significant quality ceiling on tasks requiring multi-step retrieval.

\section{Limitations}
\label{sec:limitations}

\paragraph{Benchmark--agent co-development.}
The Finance Agent was developed iteratively alongside the benchmark: dataset coverage was expanded based on evaluation results, and SME feedback on model outputs directly informed rubric refinement (Section~\ref{sec:dataset}).
The Finance Agent therefore benefited from direct exposure to benchmark signals that the general-purpose comparators did not receive, and its performance advantage should be understood partly as a product of this co-development rather than solely as an architectural difference.

\paragraph{Dataset scale.}
The benchmark comprises 251 queries.
While modest in absolute number, the dataset was designed with breadth of scenario coverage and annotation quality as primary objectives: the three task types span distinct finance workflows and data sources, and queries were curated to cover representative scenarios, edge cases, and observed failure modes with high-quality annotations.
Nonetheless, the sample size limits statistical power for fine-grained sub-segment analyses.

\paragraph{Single-turn evaluation.}
All queries are evaluated in a single-turn, single-question setting.
Multi-turn conversations, clarification exchanges, and context carryover across a session --- common in real deployments --- are outside the current scope.

\paragraph{Accuracy measurement for public-data tasks.}
Accuracy is only measured for financial obligation research, where an authoritative ERP system of record provides deterministic ground truth.
For financial entity performance research and business brief generation, strict accuracy assessment is not applied because ground truth is distributed across sources and subject to definitional variation and temporal drift (Section~\ref{sec:evaluation}); groundedness serves as the primary factual-grounding signal for those task types.

\paragraph{Single point in time.}
All inference runs were conducted at a fixed point in time; public financial data, company fundamentals, and model capabilities all evolve continuously.
Scores reported here reflect system performance as of the evaluation date and may not generalise to future model versions or updated financial data.

\paragraph{ERP QA infrastructure dependency.}
Reproducing ERP QA results requires a Microsoft Dynamics 365 Finance environment loaded with the benchmark's synthetic AP/AR data and the MCP server described in the public repository: \url{https://github.com/microsoft/FinanceBenchmark}.
Detailed setup instructions are provided in the repository, so this is not a fundamental reproducibility barrier; however, it does require a Dynamics 365 licence and provisioning effort that may limit accessibility for some researchers.
Financial entity performance research and business brief generation tasks require only web search access and are fully reproducible without additional infrastructure.

\paragraph{LLM judge bias.}
Appendix~\ref{app:judge_bias} examines cross-judge calibration and rank agreement between GPT-5.2 and Claude Opus~4.6 on a single shared inference run, finding good rank agreement and a consistent but modest calibration offset (GPT-5.2 is systematically stricter).
However, this pairwise comparison on one run does not constitute a systematic evaluation of judge bias across all providers and configurations reported in this paper.
In particular, we cannot fully rule out that GPT-5.2 as the primary judge may exhibit a self-preference bias---favouring response styles more consistent with OpenAI-generated outputs over those produced by Anthropic models---nor that any such effect is uniform across task types and metric dimensions.
The appendix analysis demonstrates that inter-provider \emph{rankings} are robust to judge choice, but absolute score levels and fine-grained margins should be interpreted with this caveat in mind.

\section{Discussion and Conclusions}
\label{sec:discussion}

\paragraph{Purpose-built agents lead across all dimensions; the gap is largest for ERP accuracy.}
The Finance Agent leads across all eight evaluation dimensions, with the largest absolute margin on accuracy for financial obligation research---a task type that requires precise, entity-resolved retrieval from a live ERP system of record, where purpose-built tool routing provides a decisive advantage over general-purpose web-search-based approaches.
When reasoning effort is raised and the timeout constraint is relaxed (the unconstrained ablation in Section~\ref{sec:timeout}), the gap between the Finance Agent and general-purpose systems narrows across most dimensions, confirming that frontier models are capable of higher-quality finance responses given sufficient time and compute budget.
However, this improvement comes at a significant latency cost (Appendix~\ref{app:latency}), and the latency effect is doubly penalising under an operational time budget: raising reasoning effort increases the wall-clock time per tool call, so applying a higher reasoning effort under the 60\,s constraint would cause the timeout to bind even more frequently than at low effort, pushing constrained performance below what low-effort runs achieve.
This interaction between reasoning effort and timeout is a critical consideration for practitioners: the 60\,s budget used in the main experiments reflects a chat-like user experience requirement, and within that regime, low reasoning effort is not just a cost-saving choice but the configuration that maximises quality for the general-purpose systems.

\paragraph{Evaluation methodology and future directions.}
Rubric-based, multi-dimensional evaluation provides a substantially richer diagnostic signal than binary pass/fail metrics, enabling practitioners to understand not just whether a system works, but along which dimensions it excels or falls short and how those margins change with deployment constraints.
Future work should extend the benchmark to multi-turn conversational settings, additional finance domains (treasury management, financial planning and analysis, regulatory reporting), and structured fact-checking approaches that could enable accuracy measurement for public-data tasks without requiring a static ground-truth corpus.

\section{Disclaimer}

\copyright~2026 Microsoft Corporation.  All rights reserved.  This document is provided "AS-IS." Information and views expressed in this document, including URL and other Internet Web site references, may change without notice. You bear the risk of using it.  Examples herein may be for illustration only.

This document does not provide you with any legal rights to any intellectual property in any Microsoft product. You may copy and use this document for your internal, reference purposes

\bibliographystyle{unsrtnat}
\bibliography{references}

\clearpage
\appendix
\section*{Appendix}

\input{appendix_judge_bias}
\input{appendix_tool_calls}
\input{appendix_latency}

\end{document}

%% file: provider_scores_table.tex
\begin{table*}[t]
\centering
\caption{Weighted scores per metric dimension and provider (95\% bootstrap CI, $n=1000$ resamples). Scores are equally weighted per plugin regardless of question count. Timeout: 60\,s (general), 300\,s (business brief). Bold = highest score in each row.}
\label{tab:provider_scores}
\resizebox{\textwidth}{!}{%
\begin{tabular}{lcccc}
\toprule
Provider & Finance Agent & Claude Opus 4.7 & Claude Haiku 4.5 & OpenAI GPT 5.5 \\
Metric &  &  &  &  \\
\midrule
Accuracy & \textbf{0.77 [0.69, 0.84]} & 0.56 [0.47, 0.66] & 0.40 [0.33, 0.47] & 0.67 [0.59, 0.75] \\
Citations & \textbf{0.79 [0.76, 0.83]} & 0.74 [0.70, 0.78] & 0.59 [0.55, 0.63] & 0.61 [0.57, 0.66] \\
Clarity & \textbf{0.84 [0.83, 0.85]} & 0.67 [0.63, 0.70] & 0.61 [0.58, 0.64] & 0.75 [0.73, 0.78] \\
Depth & \textbf{0.72 [0.70, 0.75]} & 0.55 [0.51, 0.59] & 0.44 [0.41, 0.47] & 0.63 [0.59, 0.66] \\
Groundedness & \textbf{0.76 [0.73, 0.79]} & 0.33 [0.29, 0.37] & 0.53 [0.45, 0.60] & 0.43 [0.41, 0.45] \\
Recency & \textbf{0.77 [0.74, 0.79]} & \textbf{0.77 [0.72, 0.81]} & 0.59 [0.54, 0.63] & 0.73 [0.68, 0.77] \\
Relevance & \textbf{0.86 [0.85, 0.88]} & 0.71 [0.67, 0.74] & 0.61 [0.58, 0.64] & 0.79 [0.76, 0.82] \\
Structure & \textbf{0.86 [0.85, 0.87]} & 0.74 [0.71, 0.78] & 0.65 [0.62, 0.69] & 0.79 [0.76, 0.81] \\
Overall & \textbf{0.80 [0.77, 0.82]} & 0.63 [0.59, 0.68] & 0.55 [0.51, 0.60] & 0.67 [0.64, 0.71] \\
\bottomrule
\end{tabular}
}
\end{table*}

%% file: appendix_judge_bias.tex
\section{LLM Judge Bias Analysis}
\label{app:judge_bias}

To assess whether choice of LLM judge influences benchmark conclusions, we evaluated the same set of agent responses under two judges: GPT-5.2 (primary) and Claude Opus~4.6 (validation).
The analysis was conducted on the Claude Haiku~4.5 inference run, which recorded the lowest accuracy scores of any provider in the main results; our primary concern was therefore whether the judge might systematically disadvantage this system, and if so, how much of the observed accuracy gap relative to the Finance Agent could be attributed to judge calibration rather than genuine response quality differences.  Recent research has shown that LLM judges are not always fair evaluators \citep{zheng2023judging,wang2024large}, and often prefer output produced from the own family of models \citep{panickssery2024llm}.
This yielded 25--150 matched question pairs per task-type after inner-joining on (question, task-type, segment).

\paragraph{Method.}
For each (task-type, metric) cell with at least five matched pairs we computed the per-question score delta $\Delta_i = s^{\text{GPT}}_i - s^{\text{Opus}}_i$ and applied a two-sided Wilcoxon signed-rank test to assess whether the mean offset differed significantly from zero.
P-values were corrected for multiple comparisons using the Benjamini--Hochberg false discovery rate procedure \citep{benjamini1995controlling}.
Spearman rank correlations $\rho$ were computed to assess whether judges agree on the relative quality ordering of responses, independent of any level shift.
We defined a practical significance threshold of $|\Delta| \geq 0.05$ as the minimum offset operationally meaningful for provider comparisons (a five-point swing on a 0--100 scale).

\begin{figure}[ht]
  \centering
  \includegraphics[width=\linewidth]{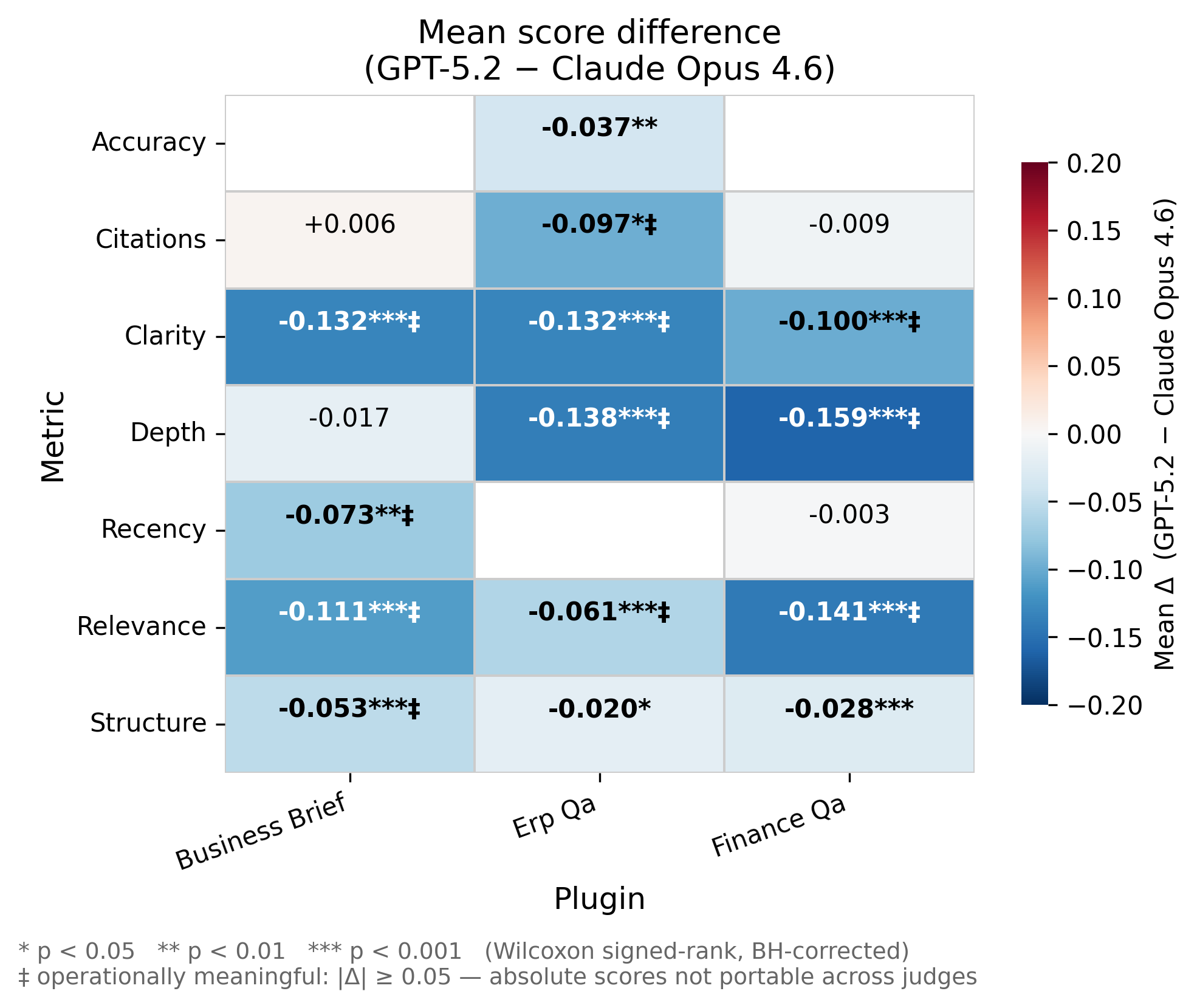}
  \caption{Mean score difference $\Delta = s^{\text{GPT-5.2}} - s^{\text{Opus~4.6}}$ per (task-type, metric) cell.
  Stars denote statistical significance after Benjamini--Hochberg correction; $\ddagger$ marks cells exceeding the practical significance threshold ($|\Delta| \geq 0.05$).
  Task-type labels: \texttt{Business Brief} = business brief generation, \texttt{Finance Qa} = financial entity performance research, \texttt{Erp Qa} = financial obligation research.}
  \label{fig:judge_delta}
\end{figure}

\paragraph{Calibration shift.}
Figure~\ref{fig:judge_delta} shows the mean score difference per (task-type, metric) cell.
GPT-5.2 is the significantly stricter judge in 14 of 18 significant cells.
Clarity shows the largest offsets across all three task types ($\Delta \approx -0.10$ to $-0.13$), consistent with GPT-5.2 applying stricter stylistic standards.
The exceptions are cells where both judges assign near-ceiling scores --- leaving no room for a systematic offset.
For business brief generation, there is no significant difference between judges for citations ($\mu \approx 0.99$--$1.00$).
There is also no significant difference between judges for financial performance entity research dimensions recency ($\mu \approx 0.941$--$0.944$) and citations ($\mu \approx 0.985$--$0.994$).
Despite relatively low scores for the depth metric for business brief generation and depth ($\mu \approx 0.367$--$0.384$), we also did not observe a significant difference between judges for these cells.

\begin{figure}[ht]
  \centering
  \includegraphics[width=\linewidth]{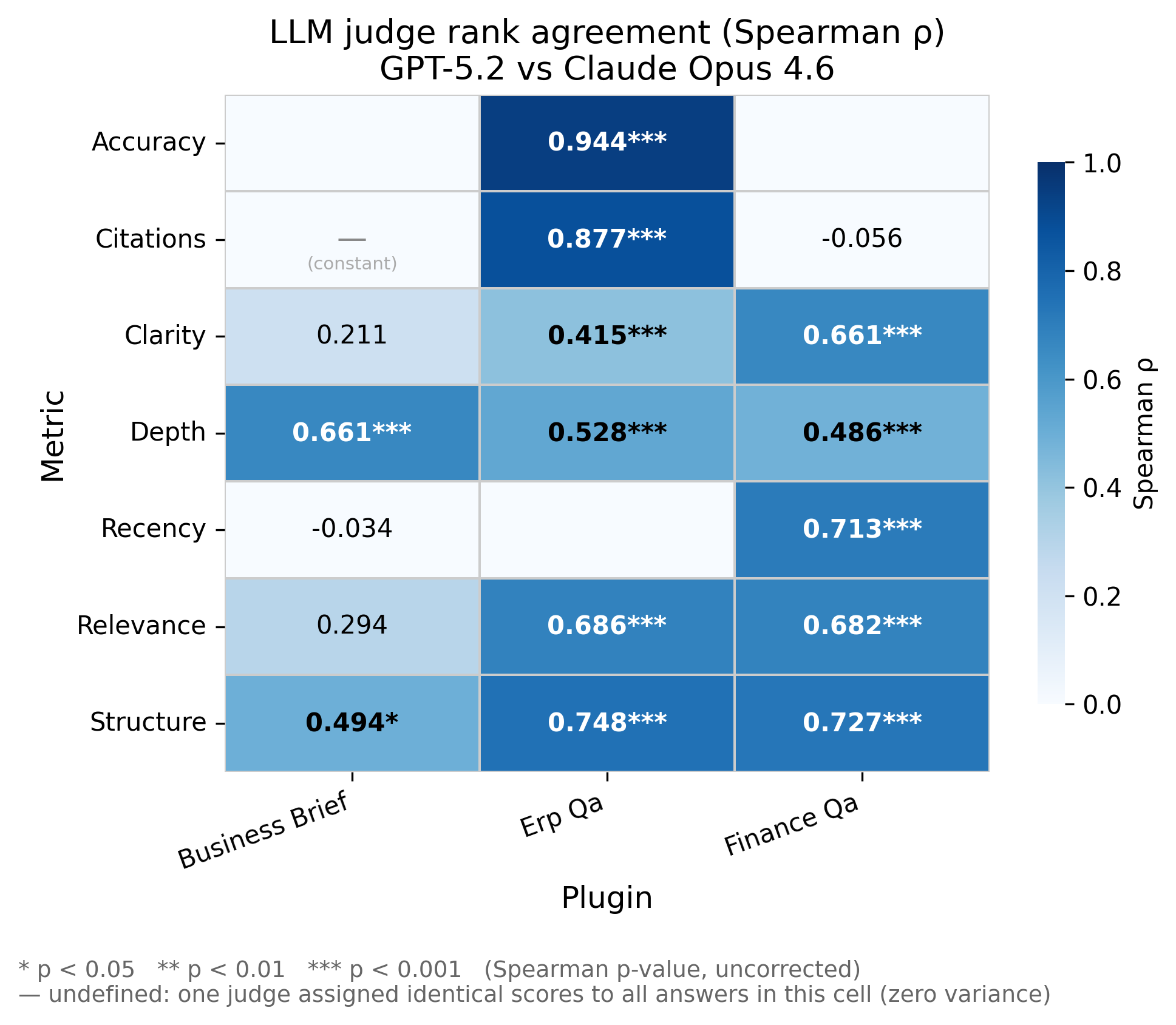}
  \caption{Spearman rank correlation $\rho$ between GPT-5.2 and Claude Opus~4.6 scores per (task-type, metric) cell.
  High $\rho$ indicates agreement on relative quality ordering independent of any level shift.
  The business brief citations cell is undefined (---) because GPT-5.2 assigned identical scores to all responses (zero variance).}
  \label{fig:judge_rho}
\end{figure}

\paragraph{Rank agreement.}
Figure~\ref{fig:judge_rho} shows Spearman $\rho$ values per cell.
ERP accuracy is the most robust cell ($\rho = 0.944$): the judges strongly agree on which responses are factually correct despite a small level offset ($\Delta = -0.037$).
The business brief citations cell reports undefined $\rho$ because GPT-5.2 assigned identical scores to all 25 responses (zero variance); this metric carries no discriminatory power for business briefs regardless of judge.

\begin{figure}[ht]
  \centering
  \includegraphics[width=\linewidth]{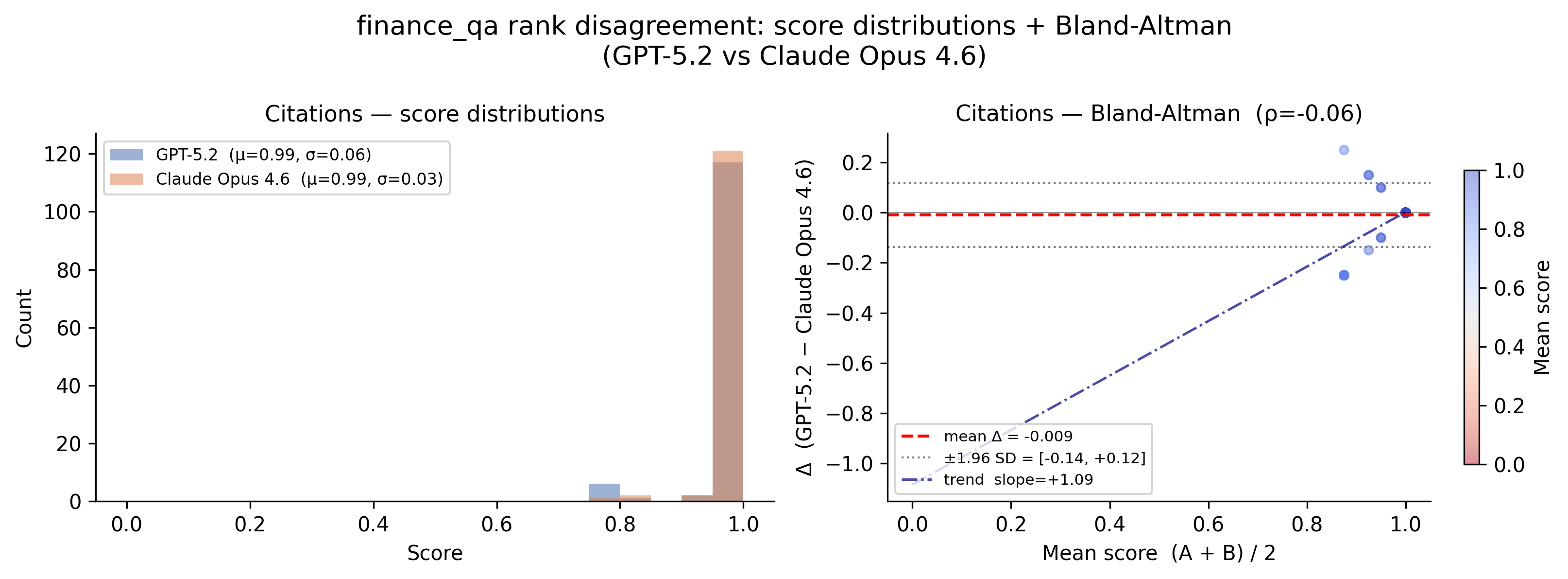}
  \caption{Score distributions (left) and Bland--Altman plots (right) for the two lowest-agreement cells: \texttt{financial entity performance research} citations.
  The positive Bland--Altman slope confirms that disagreement grows at higher score levels, consistent with a ceiling-compression artifact rather than fundamental disagreement on quality ordering.}
  \label{fig:judge_ba_finance_qa}
\end{figure}

\begin{figure}[ht]
  \centering
  \includegraphics[width=\linewidth]{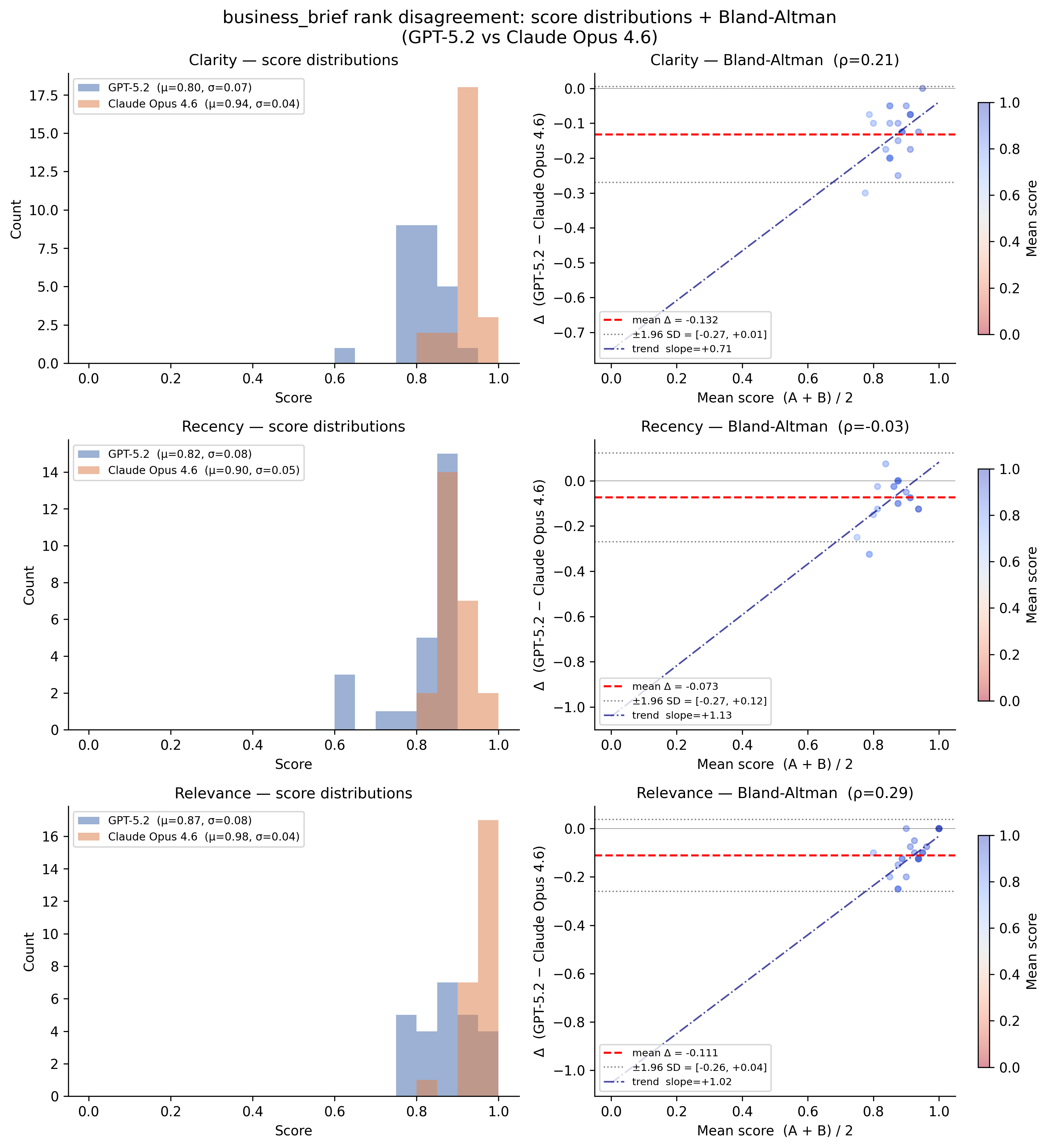}
  \caption{Score distributions (left) and Bland--Altman plots (right) for the two lowest-agreement cells: \texttt{business brief} clarity and recency.
  The positive Bland--Altman slope confirms that disagreement grows at higher score levels, consistent with a ceiling-compression artifact rather than fundamental disagreement on quality ordering.}
  \label{fig:judge_ba_business_brief}
\end{figure}

\paragraph{Rank Disagreement.}
For \emph{financial entity performance research}, citations shows the lowest rank agreement ($\rho = -0.056$). For \emph{business brief generation}, clarity, recency, and relevance show the lowest rank agreement ($\rho = -0.03$--$0.29$).
To diagnose whether this reflects genuine disagreement on quality ordering or a statistical artifact, Figure~\ref{fig:judge_ba_finance_qa} and Figure~\ref{fig:judge_ba_business_brief} show score distributions and Bland--Altman plots \citep{bland1986statistical} for the two task-types.
They show a pattern that is consistent with a ceiling-compression artifact rather than fundamental disagreement on quality ordering; the low $\rho$ values are driven by Opus's narrow score range reducing rank discrimination.

\paragraph{Implications.}

For the specific concern motivating this analysis---whether judge bias explains the accuracy gap between the Finance Agent (0.77) and Claude Haiku~4.5 (0.41) on financial obligation research---the ERP accuracy cell provides reassurance: rank agreement is very high ($\rho = 0.944$) and the mean level offset is small ($\Delta = -0.037$), meaning GPT-5.2 scores ERP accuracy approximately 0.037 lower than Opus~4.6 for Haiku responses. Even if this offset were applied as a correction, the adjusted Haiku accuracy of $\approx 0.45$ would still fall far short of the Finance Agent's 0.77, leaving the substantive gap intact. Judge bias thus accounts for only a small fraction of the observed accuracy difference.
Absolute scores are, however, not portable across evaluation runs that use different judge models; practitioners wishing to compare scores across judge runs should apply per-(task-type, metric) corrections using the offsets reported in Figure~\ref{fig:judge_delta}.

%% file: appendix_tool_calls.tex
\section{Tool-Call Distributions}
\label{app:tool_calls}

\begin{figure}[!htbp]
  \centering
  \includegraphics[width=0.64\linewidth]{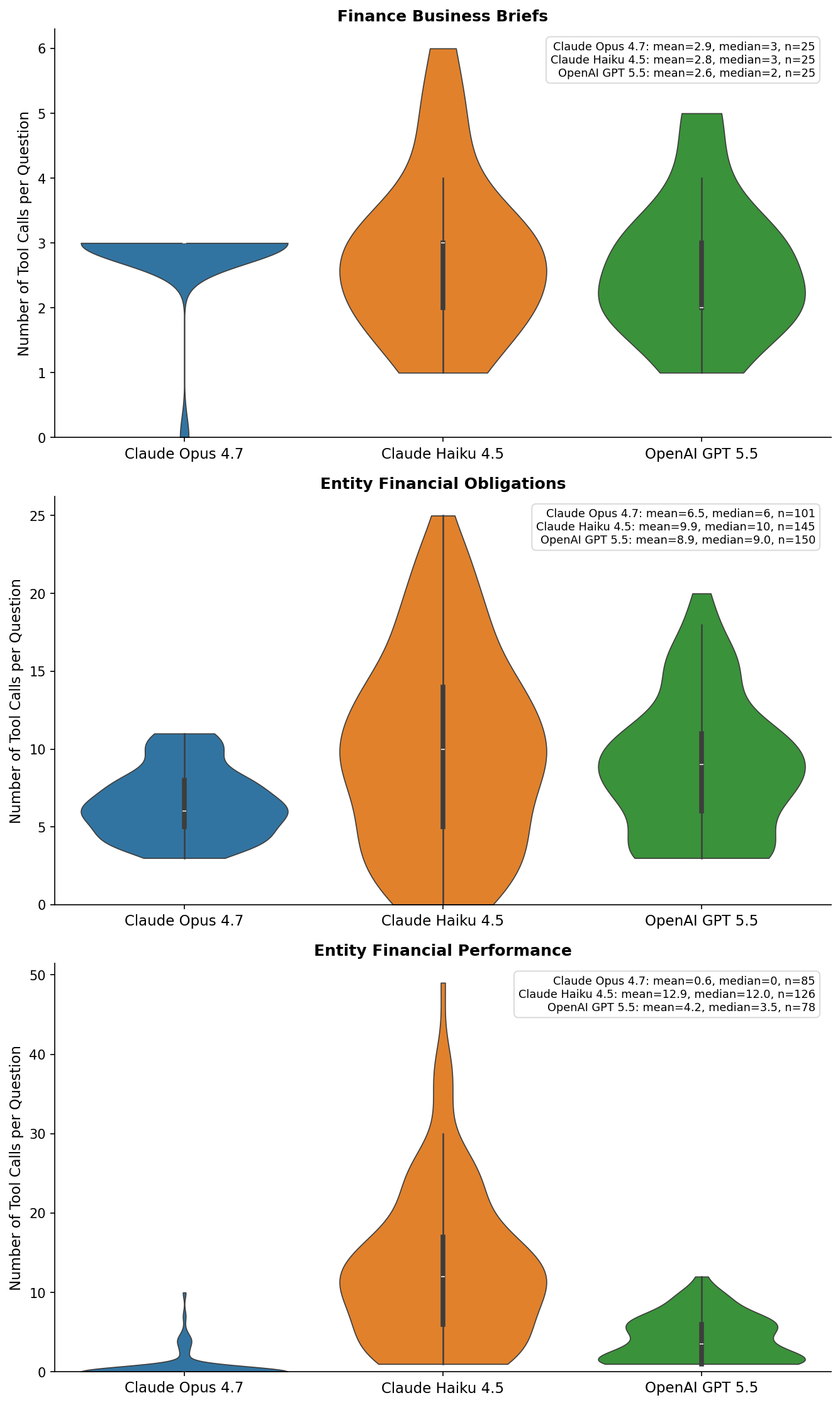}
  \caption{Distribution of tool calls per question by provider and task type.}
  \label{fig:tool_calls}
\end{figure}

Figure~\ref{fig:tool_calls} shows the distribution of tool calls per question for each provider and task type.

For financial obligation research, Claude Haiku~4.5 and GPT~5.5 issue a median of 9--10 calls per question, and Claude Opus~4.7 a median of 6; for financial entity performance research, Claude Haiku~4.5 reaches a median of 12 calls.
This higher call volume directly explains the performance degradation observed under timeout constraints (Section~\ref{sec:timeout}): agents that require many sequential retrieval steps to assemble a complete answer are disproportionately penalized by a tight time budget.

For financial entity performance research, Claude Opus~4.7 shows a median of 0 tool calls, with the distribution heavily concentrated at zero.
This is consistent with the model frequently relying on parametric knowledge rather than live retrieval for this task type under the standard reasoning configuration used in the main runs.
Interestingly, Figure \ref{fig:latency} shows that this absence of tool calls does not translate into substantially lower latency for Opus~4.7 on this task type, suggesting that the model may be spending more time internally deliberating or generating longer responses rather than issuing external calls, or that the Claude CLI did not fully capture all tool calls made by the model.

%% file: appendix_latency.tex
\section{Inference Latency Distributions}
\label{app:latency}

\begin{figure}[!htbp]
  \centering
  \includegraphics[width=.64\linewidth]{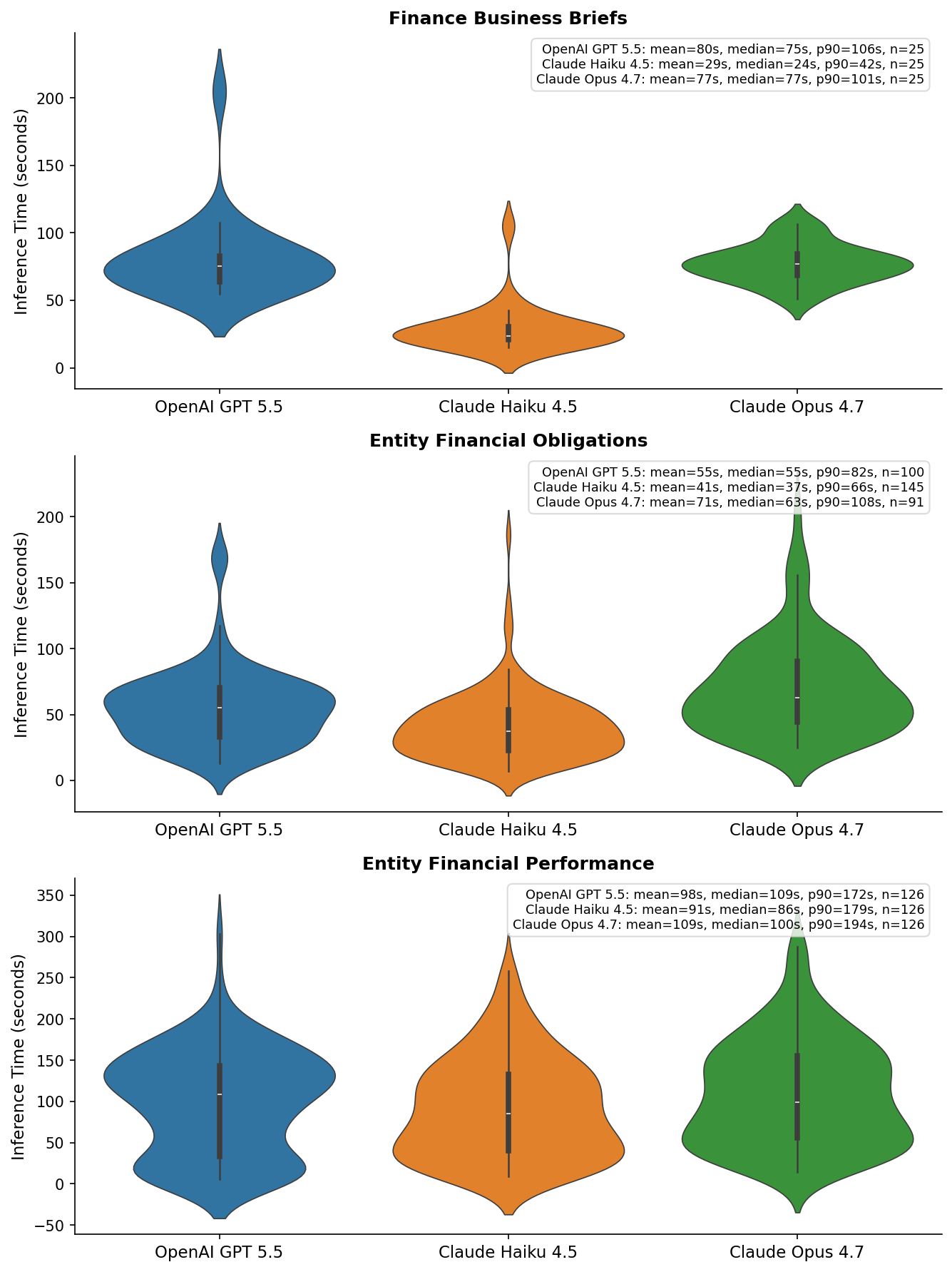}
  \caption{Inference latency distributions (violin plots with embedded box plots) for GPT~5.5, Claude Haiku~4.5, and Claude Opus~4.7 across all three task types, measured on unconstrained runs at default reasoning settings.
  Legend boxes report mean, median, p90, and sample size $n$ per provider.
  Financial entity performance research (bottom panel) has the highest latency across all providers, with p90 values of 172--194\,s, directly reflecting the high tool-call volume observed for this task type.}
  \label{fig:latency}
\end{figure}

Figure~\ref{fig:latency} shows per-question inference latency distributions for the three general-purpose providers across all task types, measured on unconstrained runs at default reasoning settings for GPT~5.5, Claude Haiku~4.5, and Claude Opus~4.7.
Finance Agent is excluded as its purpose-built pipeline has different latency characteristics not directly comparable to the general-purpose harnesses.

Claude Haiku~4.5 is substantially faster than the other two providers for business briefs (median 24\,s vs.\ 75--77\,s), consistent with its smaller model footprint.
For financial entity performance research, however, all three providers converge to similar latency ranges (mean 91--109\,s, p90 172--194\,s), reflecting the large number of sequential tool calls required by this task type regardless of model size (see Appendix~\ref{app:tool_calls}).

The latency distributions also contextualise the timeout ablation results (Section~\ref{sec:timeout}).
For financial entity performance research, the p90 latency of all three providers (172--194\,s) far exceeds the 60\,s budget, which accounts for the steep score degradation observed under that constraint.
For financial obligation research, p90 values range from 66\,s (Claude Haiku~4.5) to 108\,s (Claude Opus~4.7), meaning a substantial fraction of questions already exceed the 60\,s limit in unconstrained operation.
Business brief generation fits comfortably within the 300\,s budget for all providers (p90 $\leq$ 106\,s), consistent with the absence of meaningful score degradation under that constraint.

